\documentclass[10pt, a4paper]{article}
\usepackage{lrec2022} % this is the new LREC2022 Style
\usepackage{multibib}
\newcites{languageresource}{Language Resources}
\usepackage{graphicx}
\usepackage{tabularx}
\usepackage{soul}
\usepackage{booktabs}
\usepackage[hyphens]{url} %
\usepackage{hyperref} %
\usepackage{titlesec}
\titleformat{\section}{\normalfont\large\bfseries\center}{\thesection.}{1em}{}
\titleformat{\subsection}{\normalfont\SmallTitleFont\bfseries\raggedright}{\thesubsection.}{1em}{}
\titleformat{\subsubsection}{\normalfont\normalsize\bfseries\raggedright}{\thesubsubsection.}{1em}{}
\renewcommand\thesection{\arabic{section}}
\renewcommand\thesubsection{\thesection.\arabic{subsection}}
\renewcommand\thesubsubsection{\thesubsection.\arabic{subsubsection}}
\usepackage{epstopdf}
\usepackage[utf8]{inputenc}
\usepackage{hyperref}
\usepackage{xstring}
\usepackage{color}
\usepackage{amsmath}
\usepackage{subcaption}
\usepackage{todo}
\usepackage{url}
\usepackage[hyphenbreaks]{breakurl}

\newcommand{\bleubr}{BLEU$_{\operatorname{br}}$}
\newcommand{\bleunb}{BLEU$_{\operatorname{nb}}$}
\newcommand{\bleuem}{BLEU$_{\operatorname{em}}$}
\newcommand{\terbr}{TER$_{\operatorname{br}}$}
\newcommand{\pk}{P$_k$}
\newcommand{\eol}{\texttt{$<$eol$>$}}
\newcommand{\eob}{\texttt{$<$eob$>$}}
\newcommand{\gtag}{\texttt{$<$tag$>$}}

\title{Evaluating Subtitle Segmentation for End-to-end Generation Systems}%\\ \vspace*{.5\baselineskip} \normalfont{ The Title \ul{Must Be} Capitalised as in:\\ \vspace*{.5\baselineskip} \textbf{The Rise and Fall of Ziggy Stardust and the Spiders from Mars}}}

\name{Alina Karakanta$^{1,2}$, Fran\c{c}ois Buet$^{3}$, Mauro Cettolo$^1$, Fran\c{c}ois Yvon$^{3}$} 

\address{$^1$ Fondazione Bruno Kessler, Via Sommarive 18, Povo, Trento - Italy  \\ $^2$ University of Trento, Italy\\ $^3$ Universit\'e Paris-Saclay, CNRS, LISN, France \\
         \tt{\{akarakanta,cettolo\}@fbk.eu}, \tt{\{buet,yvon\}@limsi.fr}}

\abstract{
Subtitles appear on screen as short pieces of text, segmented based on formal constraints (length) and syntactic/semantic criteria. Subtitle segmentation can be evaluated with sequence segmentation metrics against a human reference. However, standard segmentation metrics cannot be applied when systems generate outputs different than the reference, e.g. with end-to-end subtitling systems. In this paper, we study ways to conduct reference-based evaluations of segmentation accuracy irrespective of the textual content. We first conduct a systematic analysis of existing metrics for evaluating subtitle segmentation. We then introduce $Sigma$, a new Subtitle Segmentation Score derived from an approximate upper-bound of BLEU on segmentation boundaries, which allows us to disentangle the effect of good segmentation from text quality. To compare $Sigma$ with existing  metrics, we further propose a boundary projection method from imperfect hypotheses to the true reference. Results show that all metrics are able to reward high quality output but for similar outputs system ranking depends on each metric's sensitivity to error type. Our thorough analyses suggest $Sigma$ is a promising segmentation candidate but its reliability over other segmentation metrics remains to be validated through correlations with human judgements. %we review existing segmentation metrics and their suitability for evaluating subtitle segmentation. We perform an analysis which shows the possibility to disentangle the impact of text changes and segmentation quality in a BLEU-based metric. Lastly, we propose a boundary projection method based on hypothesis-reference alignment, which allows applying standard metrics when the content of the output does not match that of the reference. 
\\ \newline  }

\begin{document}

\maketitleabstract
%\color{red} check on diacritics in utf8 :  à - â - ä - é - è - ê - ë - ï - î - ô - ö - ù - û - ü - ÿ - ç %\color{black}
% ALl values have been changed KC20210901

% 4-8 pages, unlimited refs
\section{Introduction \label{sec:introduction}}
Accessibility of audiovisual content  has become a legal obligation for major TV channels in many countries \cite{eu-av-10} and is also a strong suggestion for uploaded web content \cite{eu-web-16}. Subtitles are a means for providing accessibility services, either with intralingual closed captioning for the deaf and hard-of-hearing or with interlingual subtitling in various languages for persons without knowledge of the source language speech. Subtitles are also useful for online talks and educational content and facilitate the comprehension of speech by language learners. Automatising the generation of subtitles has been a long-standing issue \cite{Piperidis04multimodal,Melero06automatic,Volk10machine}, and is nowadays more and more often performed with neural models trained end-to-end \cite{Lakew19controlling,Liu20adapting}. 

Automatic generation of subtitles is a difficult task, since subtitles should not only reflect the spoken content, but should also satisfy multiple formal requirements related to their position on screen, the text length, size and colour, their display duration and synchronization with speech, etc. Additionally, a good segmentation of the transcribed or translated text into subtitles must satisfy syntactic and semantic constraints, since a segmentation which respects linguistic units has been shown to facilitate comprehension and to lead to more readable subtitles \cite{perego2008break,Rajendran2013chunking}. Subtitling must thus go well beyond the automatic transcription and translation of the soundtrack, and complete automation would typically require additional processing modules such as text simplification and segmentation into readable chunks, speaker diarization and sound event detection.

In this work, we study ways to evaluate the quality of the output segmentations delivered by end-to-end subtitling systems. Contrary to pipeline systems, which typically contain an independent segmentation module that can be evaluated as a standalone component by simply measuring its ability to reproduce a reference segmentation of a reference text (\textit{perfect text}), end-to-end systems directly output a segmented text, which may not correspond to the reference subtitle (\textit{imperfect text}). Separating text from segmentation errors thus becomes an issue. Recent proposals to address this problem notably include metrics such as \bleubr{} and \terbr{} \cite{Karakanta20is42}, \textit{inter alia}, which include segmentation tags in the computation of the overall output quality. However, their ability to single out segmentation errors remains unclear. In this work, we perform a systematic assessment of subtitle segmentation metrics, with the aim to better understand their behaviour in relation to textual errors. 

% The need to evaluate subtitle segmentation comes both when segmentation is performed by an independent component in an NLP pipeline, and with more integrated (e.g.\ end-to-end) systems, which simultaneously generate textual content (simplification / translation) and segmentation boundaries. %Problem formulation

Our contributions can be summarised as follows:
\begin{itemize}
\item A comparison of existing sequence segmentation metrics to evaluate subtitle quality in the ideal situation of a \textit{perfect} textual content, exactly matching the reference (Sec.~\ref{exp1});
    \item A new score $Sigma$, derived from an estimation of an upper bound of \bleubr{}, which isolates the segmentation signal irrespective of text quality, for \textit{imperfect} texts (Sec.~\ref{exp2});
    \item A boundary projection method which maps the subtitle breaks from hypothesis to the reference and allows for applying the standard segmentation metrics even for \textit{imperfect} texts (Sec.~\ref{exp3});
    \item EvalSub: A tool for computing reference-based segmentation scores for automatic subtitles.\footnote{Our code to replicate the experiments is available at \url{https://github.com/fyvo/EvalSubtitle}.}
\end{itemize}

\section{Generating and evaluating subtitle segmentation \label{sec:segmenting}}
\subsection{Problem statement} \label{subsec:problem}
\begin{figure}
    \centering
    \includegraphics[width=0.35\textwidth]{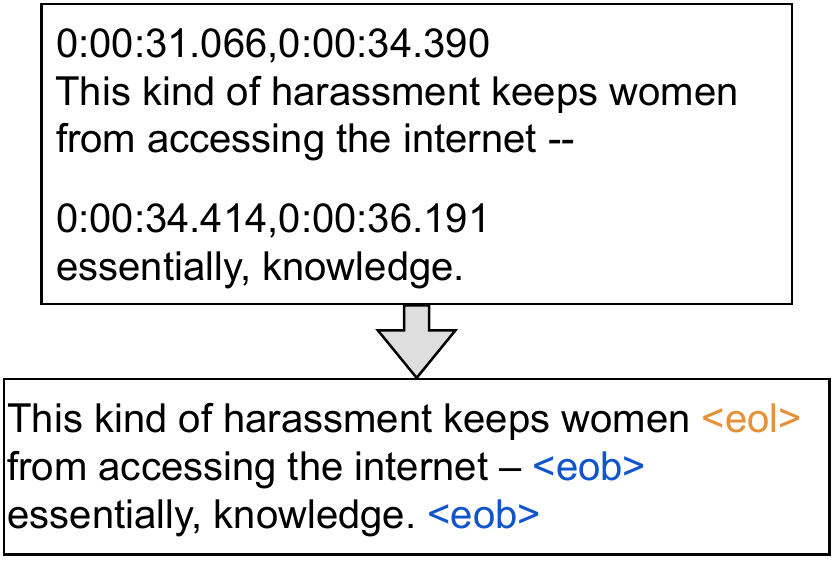}
    \caption{Example of the \eol{}, \eob{} segmentation notation for two subtitle blocks.}
    \label{fig:example}
\end{figure}

In this work, we only focus on the evaluation of segmentation, and consider that the system's output is composed of text interspersed with segmentation symbols. We further assume that there are two types of symbols: \eol{}, which indicates a change of line within the same screen, and \eob{}, which indicates the end of a subtitle block and a subsequent change of screen. Figure~\ref{fig:example} displays an example of two subtitle blocks using these notations. According to subtitling guidelines \cite{bbc,Netflix,ted}, there should be no more than two lines on the same screen, and each line should contain about 40 characters, with variations depending on the language and audience. To ease readability, line and subtitle breaks %, and even more so, screen changes, 
should be positioned so as to preserve the syntactic and semantic units as much as possible \cite{Carroll-Ivarsson-98}. In addition, the display duration of each subtitle should vary according to the number of characters on screen (reading speed), while keeping in sync with the spoken content.

% desiderata for metric? hypotheses?
Matching these constraints is a challenging task, and human subtitlers often have to compromise one constraint over the other. Moreover, the decision of where to segment depends not only on the syntax/semantics of the text \cite[p. 172]{cintas-07-sub} but also on multimodal factors such as speaker changes, speech pauses and shot changes. %For these reasons there is a lot of variability in subtitling: there is still an open debate on whether one longer line is easier to process compared to two shorter lines \cite{perego-line-08} and this decision often depends on the type of content \cite{Szarkowska-onlinesurvey-16}. 
This means that a lot of technical, linguistic and extra-linguistic expertise goes into the segmentation process, and that there is a lot to learn from corpora of actual subtitles, be they intralingual or interlingual. %For this reason, reference-based evaluation allows for taking into account this expertise by comparing the similarity of the automatic segmentation with segmentations based on human decisions, without risking to ignore important multimodal and linguistic aspects leading to these decisions.

\subsection{Metrics for segmentation}
As for evaluating automatic segmentation, there are two possible approaches: one is to separately evaluate how well each constraint is matched, then derive an aggregate score; the other is to try to reproduce human reference segmentations. Both have their strengths and weaknesses: the former is difficult due to the need to perform a syntactic and semantic analysis of the subtitles, and to correctly weight the importance of each of the constraints listed above; the latter is tricky because standard string comparison metrics are not appropriate for subtitles. For instance, minor changes such as adding one extra \eol{} can yield an invalid display with three lines%; adding an extra \eob{} will lead to shorter %but faster subtitles% since increasing the number of subtitles will accordingly increase the reading speed
. Conversely, missing one \eol{} can make the current line overly long to even fit the screen. As for the position of the boundaries, it may also happen that moving a break three words ahead can better match the syntactic constraint than moving it just one word ahead.

% THIS IS NOT THE CASE in SUBTITLING: Near misses occur frequently in segmentation—although manual coders often agree upon the bulk of where segment lie, they frequently disagree upon the exact position of boundaries (Artstein and Poesio, 2008, p. 40). 
We focus here on \emph{reference-based metrics} and further require that a good segmentation metric should:%\todo[FY]{To be discussed and completed}
\begin{itemize}
\item account for different types of boundaries;
\item accommodate scenarios where multiple human references are available;
\item handle content differences with the reference;
\item disentangle the effect of a poor content from that of a poor segmentation; 
\item realise a fair balance between all the formal and structural constraints.
\end{itemize}

We thus conduct three experiments where we 1) analyse standard segmentation metrics for evaluating subtitling segmentation and discuss how well each of them accommodates the above criteria for perfect texts (Section~\ref{exp1}), 2) propose \textit{Sigma}, a new score derived from \bleubr{}, which allows for disentangling the effect of text quality from segmentation (Section~\ref{exp2}) and 3) compare all metrics on real subtitling tasks for outputs generated by end-to-end neural machine translation and speech translation systems (Section~\ref{exp3}).

\section{Experimental setting} 

\subsection{Metric sensitivity/robustness} \label{exp1}
In the first experiment, we investigate the behaviour of standard segmentation metrics and metrics previously used in the evaluation of subtitle segmentation for perfect texts in an artificial environment where we control the degree of drop in segmentation quality. The metrics are the following:

\begin{itemize}
    \item \textbf{Precision, recall and F1} \cite{alvarez-et-al-2016}:
\textbf{Precision} is defined as the proportion of boundaries in the hypothesis that agree with the reference boundaries over the total number of hypothesis boundaries, while \textbf{recall} is the number of correct boundaries divided by the reference boundaries. \textbf{F1} is the harmonic mean of precision and recall. %These metrics cannot account for different boundary types (end of line vs end of block) and equally penalise near and full misses. %TODO: how do we compute it here
 
 \item \textbf{Window-based metrics}: \textbf{\pk{}} \cite{beeferman-berger-99-pk} assigns penalties for each moving window if ends are detected to be in different segments between reference and hypothesis, while 
 %\pk{} measures the probability that two units $k$ steps apart are incorrectly identified as belonging to different segments. It computes penalties via a moving window of length $k$, where $k$ is set to half of the average true segment size. For each window, the algorithm determines whether boundaries are detected to be in different segments of the reference and hypothesis segmentations and in this case assigns a penalty of 1. The final score is computed as the total penalty normalised by the number of windows. 
 \textbf{WindowDiff}
\cite{pevzner-hearst-2002-critique} assigns a penalty if the number of boundaries in each window is different for reference and hypothesis.
 %For each moving window, WindowDiff compares the number of boundaries in the reference that fall in the window to the number of boundaries in the hypothesis. It assigns a penalty if the number of boundaries in each window is different for reference and hypothesis. Both \pk{} and WindowDiff differently penalise near and full missed but still cannot account for different boundary types.
 
 \item \textbf{Edit distance-based metrics}:
 \textbf{Segmentation similarity}
\cite{fournier-inkpen-2012-segmentation} computes the proportion of boundaries that are not transformed when comparing segmentations using edit distance as a penalty function. \textbf{Boundary similarity}
\cite{fournier-2013-evaluating} is an adaptation of segment similarity, where different weights are applied for each edit type.  
In \textbf{\terbr{}} \cite{Karakanta20is42} all words except boundary symbols in each hypothesis-reference pair are masked and TER \cite{Snover06study} is computed over the masked sequences.

\item \textbf{\bleubr{}} \cite{Karakanta20is42}: 
BLEU computed on text containing subtitle boundaries as special symbols. It has often been reported together with \bleunb{} (no boundaries), computed over the hypothesis-output without the boundary symbols. %Since each boundary is a different token, the metric can consider different types of breaks. However, the relationship between BLEU with and without boundaries has not been explored.
\end{itemize}

%It computes the proportion of boundaries that are not transformed when comparing segmentations using edit distance. Edit distance is used as a penalty function and penalties are scaled by segmentation size. Penalties are subtracted for each edit operation, normalised by the total number of potential boundaries in the hypothesis. %S is a symmetric function that quantifies the similarity between two segmentations as a percentage, and applies to any granularity or segmentation unit (e.g., paragraphs, sentences, clauses, etc.).  full misses as the addition/deletion of a boundary, and near misses as n-wise transpositions. 
%Boundary similarity is an adaptation of segment similarity, where different weights are applied for each edit type (additions, deletions, transpositions, substitutions). The score is calculated as one minus the incorrectness of each boundary pair over the total number of boundary pairs. %correctness score for each boundary pair/decision and then using the mean of this score as a normalization of boundary edit distance. Edit-based metrics can account for different boundary types and the distance is considered(?)

To investigate the sensitivity/robustness of the segmentation metrics in subtitling tasks, we perform changes in the reference segmentation in a controlled way. Specifically, we apply the following operations randomly on the reference segmentation: 
1) shift, where a boundary is shifted 1, 2 or 3 positions to its left/right, 
2) addition, where a new boundary is added to the segment, 3) deletion, where a boundary is deleted from the segment, and 4) replacement, where a boundary is substituted with the other boundary type, e.g. \eol{} substituted with \eob{}. For each operation type, we gradually increase the percentage of boundaries affected by the operation (20\%, 40\%, 60\% 80\% and 100\%). For example, shift.1.20 corresponds to 20\% of the reference boundaries shifted by one position, while delete.80 means that 80\% of the reference boundaries are deleted. Additions are made with respect to the number of boundaries and not to the number of possible insertion positions (spaces); that is, add.100 doubles the number of boundaries. Finally, metrics are computed between the modified test set and the true reference.

\subsection{\bleubr{}: Content vs.\ segmentation } \label{exp2}

\begin{figure*}[t]
  \centering
  \begin{tabular}{l p{0.9\textwidth}}
    \toprule 
    REF &  the car has just left Paris \eol{} for its destination London \eob{} where it will arrive next Sunday \eol{} if all goes well . \eob{} \\
    HYP & the car has just left Paris \eol{} for his destination : London \eob{} where he arrives \eol{} next Sunday if \eol{} all goes well . \eob{} \\
    \bottomrule
  \end{tabular}

  \begin{tabular}{p{0.99\textwidth}} \\
    The bigram ``\textcolor{blue}{left Paris}'' and trigram ``\textcolor{blue}{all goes well}'' are counted both by \bleunb{} and \bleubr{}; the bigram ``\textcolor{blue}{London \eob{}}'' and the trigram ``\textcolor{blue}{Paris \eol{} for}'' are counted by \bleubr{} but not by \bleunb{}; conversely the bigram ``if all'' and trigram ``arrive next Sunday'' are counted by \bleunb{}, but not by \bleubr{}.
   \\
  \end{tabular}
  
%  \vspace{\baselineskip}
  
  \caption{Comparing \bleunb{} and \bleubr{}}
  \label{fig:bleubr}
\end{figure*}

In the second experiment, we explore whether \bleubr{} really captures segmentation quality. BLEU without (\bleunb{}) and with (\bleubr{}) boundaries have sometimes been reported together (e.g. in \cite{Karakanta20is42} and \cite{Buet21toward}), the motivation being that \bleunb{} should evaluate the content, and \bleubr{} the segmentation. Yet, the relationship between these two scores suggests that this interpretation may be oversimplistic, motivating deeper analyses.

\bleubr{} is computed on longer sequences than \bleunb{}, which means more $n$-grams to match. Since predicting the right number and type of segmentation tags is generally easier than predicting the actual words, \bleubr{} usually has a higher unigram precision, which can, in turn, impact the higher-order precision scores.\footnote{Almost all our simulations have a higher unigram precision for \bleubr{} than for \bleunb{}.}
This suggests that the absolute or relative difference between the two scores cannot be a proper signal of segmentation quality alone: interpreting \bleubr{} $>$ \bleunb{} as a sign of good segmentation may be correct, but the intensity of this signal cannot be realistically assessed from these two measures alone.
How to compare \bleunb{} and \bleubr{}? When \bleunb{}$=$100, as with perfect texts, \bleubr{} cannot be greater; in that case, decreases of \bleubr{} directly reflect segmentation errors. With imperfect texts though, the more \bleunb{} goes down, the easier it is to observe \bleubr{} values greater than \bleunb{}. % (see Figure~\ref{fig:bleu2}). 

% Note that we can also have \bleunb{} $<$ \bleubr{}.

Differences between \bleubr{} and \bleunb{}, as shown in Figure~\ref{fig:bleubr}, result from matches obtained for the $n$-grams containing a segmentation tag (henceforth \emph{$n$-tagrams}). A first possible way to disentangle the effects of the segmentation would thus be to separately compute two scores: \bleunb{} and \bleuem{},\footnote{This approach is investigated by \'Elise Michon, personal communication with the authors.} where \bleuem{} only measures precision scores with respect to $n$-tagrams. Unigram precision only counts segmentation tags, bigram precision counts 2-tagrams such as ``w \gtag{}'' or ``\gtag{} w'' etc. However, \bleuem{} remains highly correlated with \bleunb{}. This is because $n$-tagram matches for \bleuem{} directly depend on the precision of $(n\!-\!1)$-grams for \bleunb{}. For instance, ``w$_1$ w$_2$ \gtag'' can only be correct if ``w$_1$ w$_2$'' is also a match, implying that the $n$-gram scores define an upper bound on the possible $n+1$-tagram matches.

We therefore discard \bleuem{} and consider instead an upper bound of \bleubr{}, denoted \bleubr$^+$ and computed as follows.
We denote $p_1, p_2, p_3, p_4$ respectively as the 1g, 2g, 3g and 4g modified precisions computed by \bleunb{}, $\alpha$ the ratio of the number of boundaries to the number of word tokens, and $p^{\prime}_1, \dots{}, p^{\prime}_4$ the corresponding precisions for \bleubr{}. Under the assumption that boundaries are mostly correct, the expected number of correct unigrams in a text of $l$ words augmented with boundaries is just $p_1 \times l + \alpha \times{} l$, yielding $p^{\prime}_1 = \frac{p_1 + \alpha}{1 + \alpha}$. For higher order n-grams, the exact computation is more involved, but a simple upper bound is the following:
$$
p^{\prime}_{n} \le \frac{(1 - (n-1) \alpha) \times p_{n} + n \alpha \times p_{n-1} }{1+\alpha}.
$$
This holds because we assume that:
%(a) all correct $n$-grams matching for \bleunb{} also match for \bleubr{};
%(a) all correct $n$-grams matching for \bleunb{} also match for \bleubr{};
%(b) each $n-1$-gram matched for \bleunb{} yields (at most) one new n-tagram\footnote{In theory it might create two: one with the tag on the right, one with a tag on the left, but this would imply very short lines which rarely appears in our references.};
(a) each boundary is part of $n$ $n$-tagrams, within which it is surrounded by regular tokens;\footnote{In theory there might be several breaks in one $n$-tagram, but this would imply very short lines and rarely appears in our references.} and
(b) the text is sufficiently long so that we can make the approximation $\frac{l}{l+1} \approx{} 1$.
We readily derive an upper bound \bleubr$^+$ of \bleubr{} that we can compute from \bleunb{}. This value can be used as a proxy to the best achievable \bleubr{} score for a given \bleunb{}. We thus denote our new score $Sigma$ ($S$) as:
\begin{equation}
S = \frac{\text{\bleubr}}{\text{\bleubr}^+} \label{eq:sigma}
\end{equation}
Values close to $100$ should signal a good segmentation, while values close to $0$ a bad segmentation, \emph{irrespective of the value of} \bleunb.

We empirically investigate the above assumptions in two steps. We first explore the relation between \bleubr{} and \bleunb{} for imperfect system outputs when the segmentation remains constant. To simulate these outputs, we insert noise in the reference text, without affecting the type or position of boundaries. The noising process consists of applying a mix of edit operations (insertions, deletions, substitutions, in equal shares) corresponding to a certain percentage of the number of tokens (from 0 to 90, with a step of 10). We then move to the case when textual errors in imperfect texts are combined with segmentation errors. We apply the segmentation changes (mix of operations on the boundaries, following the same procedure as for the words) to the noisy references generated in the first step and compare the behaviour of our new score $Sigma$ compared to \bleubr{} for different values of \bleunb{}.
 
\subsection{Boundary projection} \label{exp3}
\begin{figure}[!ht]
  \center
  \includegraphics[width=\columnwidth]{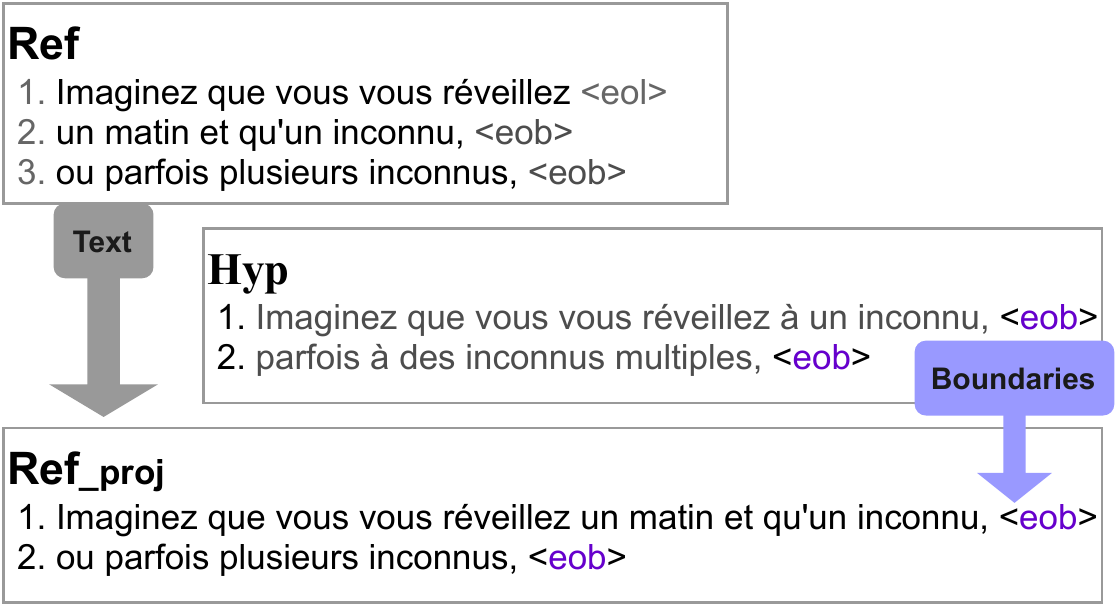}
  \caption{Projection of boundaries from hypothesis to reference based on subtitle alignment.\label{fig:alignment}}
\end{figure}

In the third experiment, we move from the scenario of controlled text and segmentation changes to investigating the usefulness of $Sigma$ for evaluating the output of real end-to-end subtitling systems. To this aim, we compare $Sigma$ to \bleubr{} and \terbr{}, as well as to the scores obtained by standard segmentation metrics. To overcome the fact that standard metrics cannot be computed on imperfect texts, we apply a boundary projection method based on reference-hypothesis alignment, illustrated in Figure~\ref{fig:alignment}. Given a reference-hypothesis segment pair $\operatorname{Ref}(1, .., i)$ and $\operatorname{Hyp}(1, .., j)$, where $i$ and $j$ are respectively the number of subtitles in the reference and hypothesis segment, we split the reference and hypothesis at the subtitle boundaries, such that each subtitle (or subtitle line) is one segment. %appears on a different line. 
Then, the reference subtitles are aligned to the hypothesis subtitles using the MWER algorithm \cite{matusov-etal-2005-evaluating}. After this process we obtain a new reference $\operatorname{Ref}_{proj}(1, .., j)$, containing the text of the true reference but with the boundaries projected from the hypothesis. Since $\operatorname{Hyp}$  and $\operatorname{Ref}_{proj}$ have the same number of subtitles, the boundaries of the hypothesis are simply copied in the $\operatorname{Ref}_{proj}$. Projecting the boundaries from the hypothesis to the reference allows us to compute standard segmentation metrics between the projected reference $\operatorname{Ref}_{proj}$ and the true reference $\operatorname{Ref}$, as in Experiment~\ref{exp1}.

For comparison with previous work, we apply the boundary projection on the outputs of the 4 systems of \newcite{Karakanta20is42} for En$\xrightarrow{}$Fr. The systems are a neural MT system (\texttt{NMT}), a cascade speech translation system (\texttt{Cas}), and two end-to-end ST systems: \texttt{e2e$_{base}$} for an ST system trained only on MuST-Cinema and \texttt{e2e$_{pt}$} for a ST system pretrained on large amounts of ST data and fine-tuned on MuST-Cinema. We then compute the segmentation metrics and discuss how the ranking of system outputs based on $Sigma$ scores differs wrt. 1) standard segmentation metrics applied on the projected reference and 2) the \bleubr{} and \terbr{} computed between the output (without projection) and the true reference. % as reported in \newcite{Karakanta20is42}. %2) whether and which metrics agree on their ranking of the system outputs and 2) how the \bleubr{} and \terbr{} scores differ when computed on the original system output compared to our projected reference.

\begin{figure*}[t]
\centering
	\begin{subfigure}[b]{0.4\textwidth}
        \includegraphics[width=\textwidth]{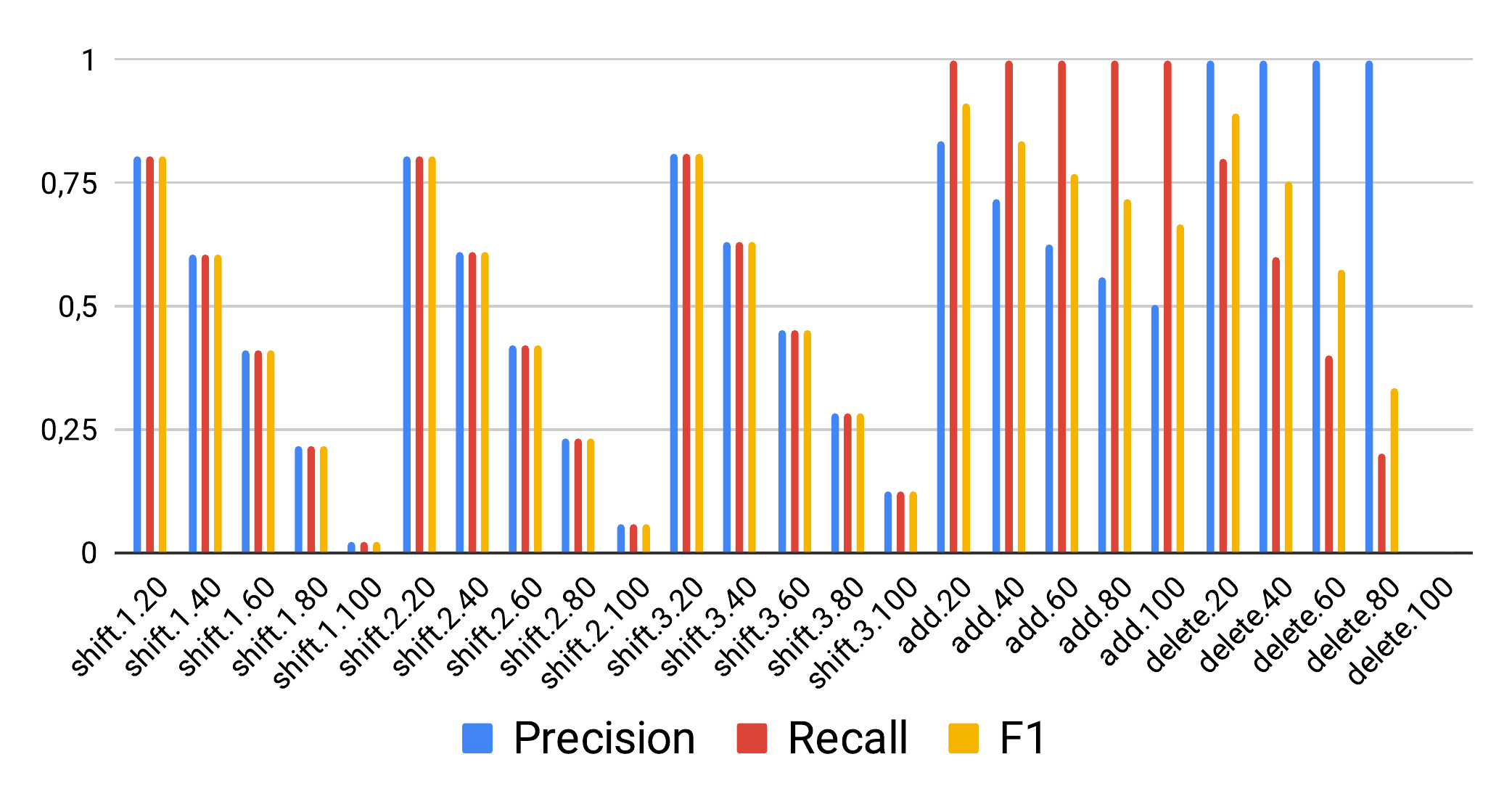}
        %\caption{...} \label{fig:a}
    \end{subfigure}
    \begin{subfigure}[b]{0.4\textwidth}
        \includegraphics[width=\textwidth]{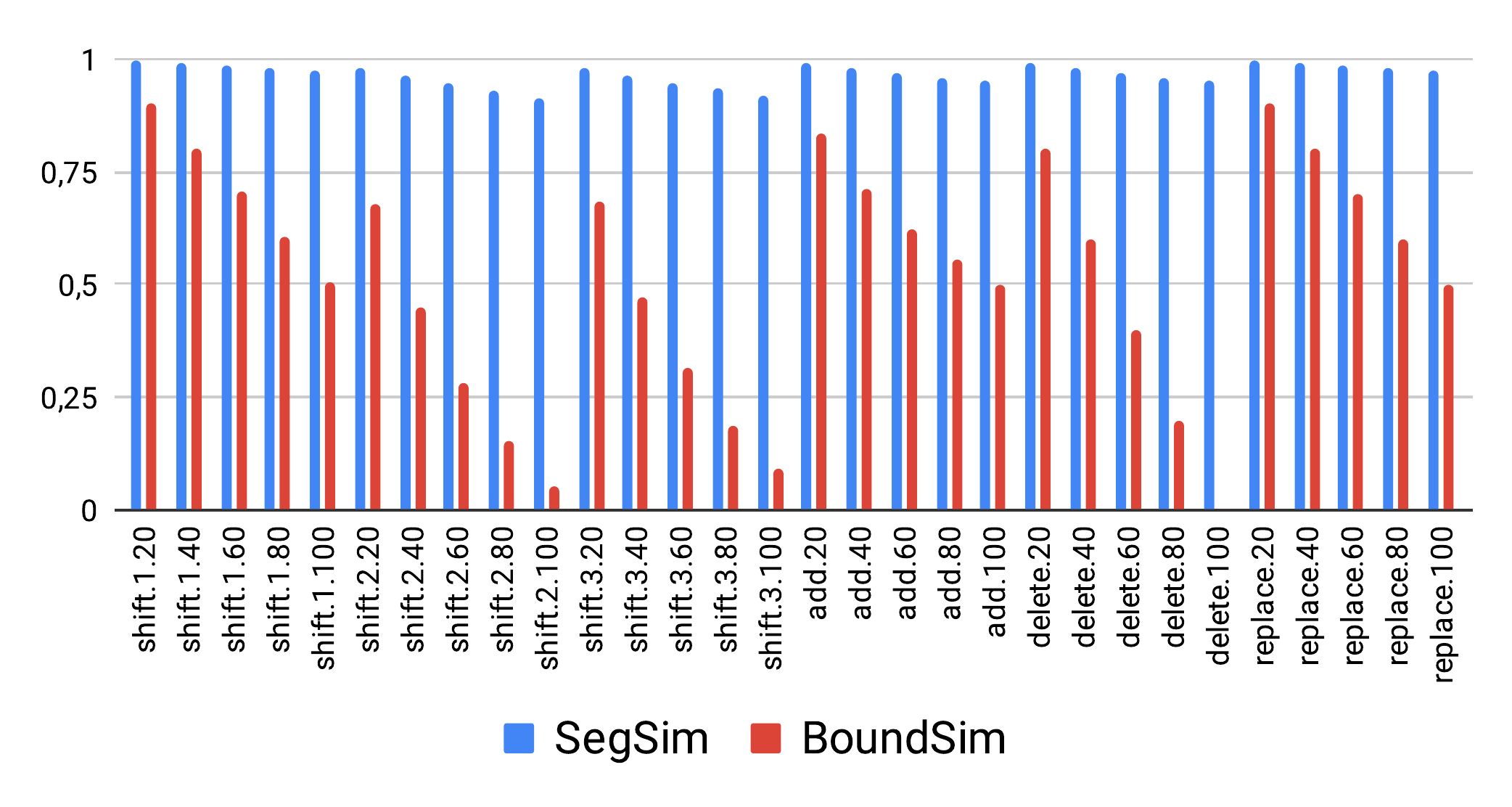}
        %\caption{...} \label{fig:b}
    \end{subfigure}
    \begin{subfigure}[b]{0.32\textwidth}
        \includegraphics[width=\textwidth]{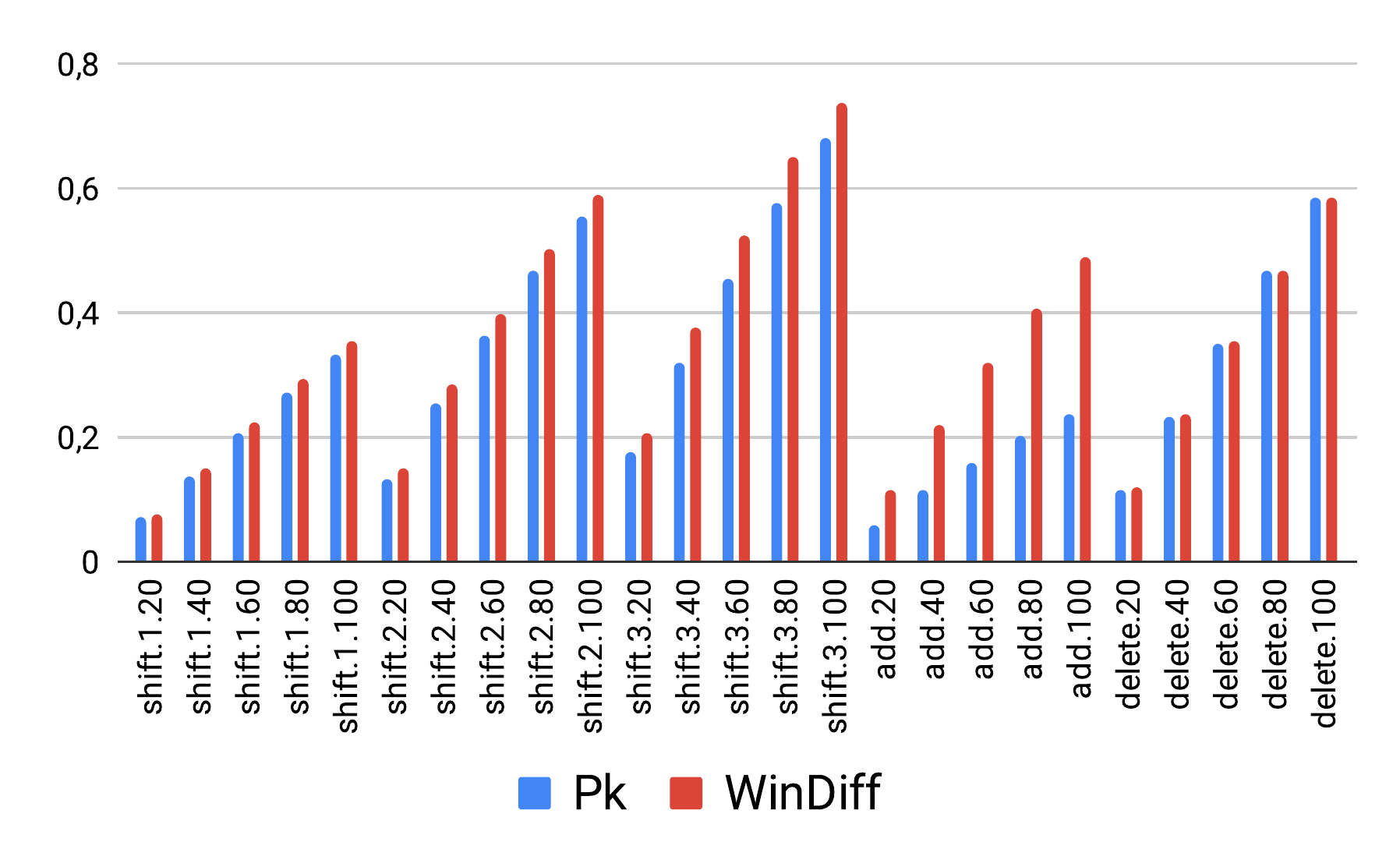}
        %\caption{...} \label{fig:c}
    \end{subfigure}
    \begin{subfigure}[b]{0.32\textwidth}
        \includegraphics[width=\textwidth]{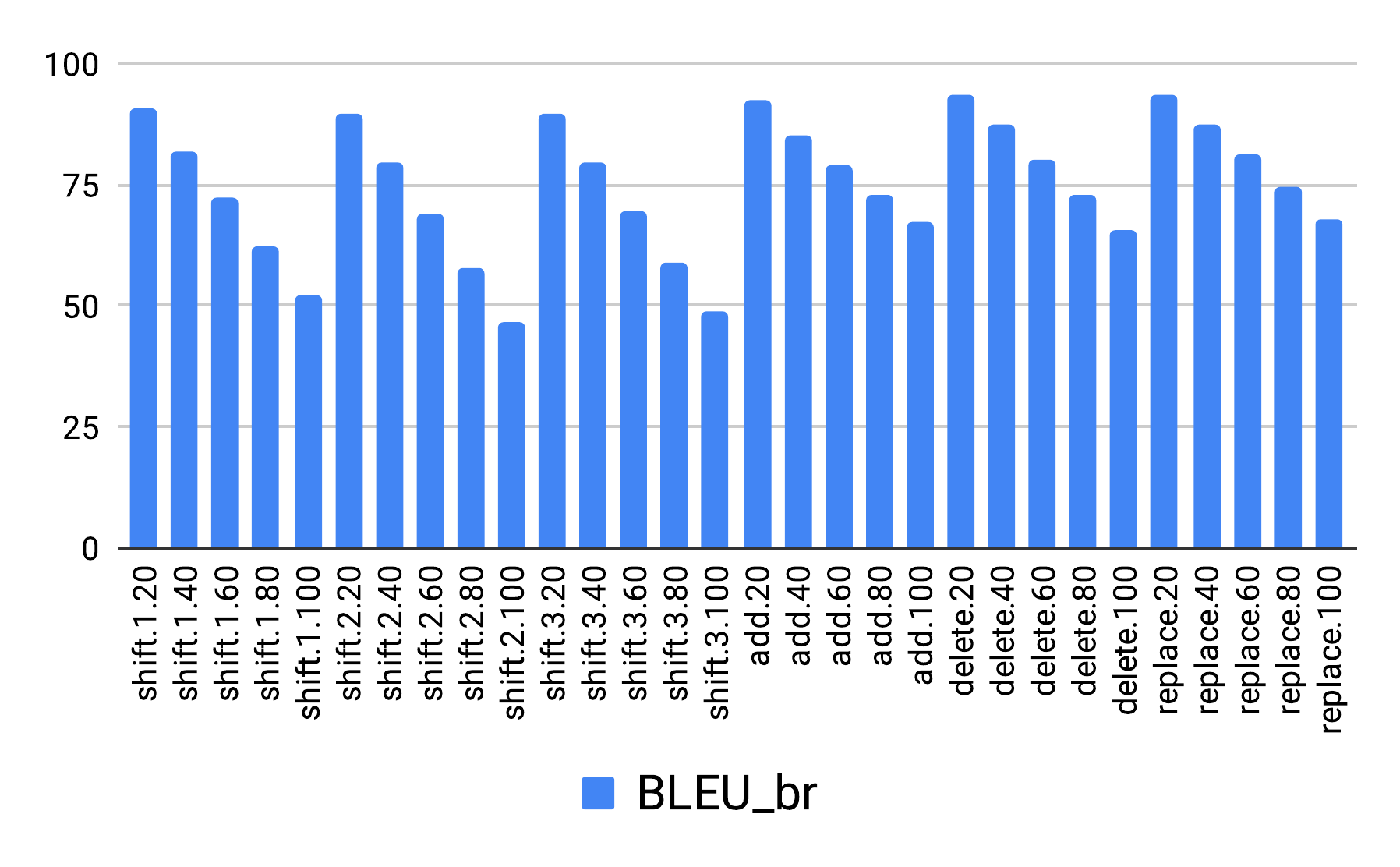}
        %\caption{...} \label{fig:d}
    \end{subfigure}
    \begin{subfigure}[b]{0.32\textwidth}
        \includegraphics[width=\textwidth]{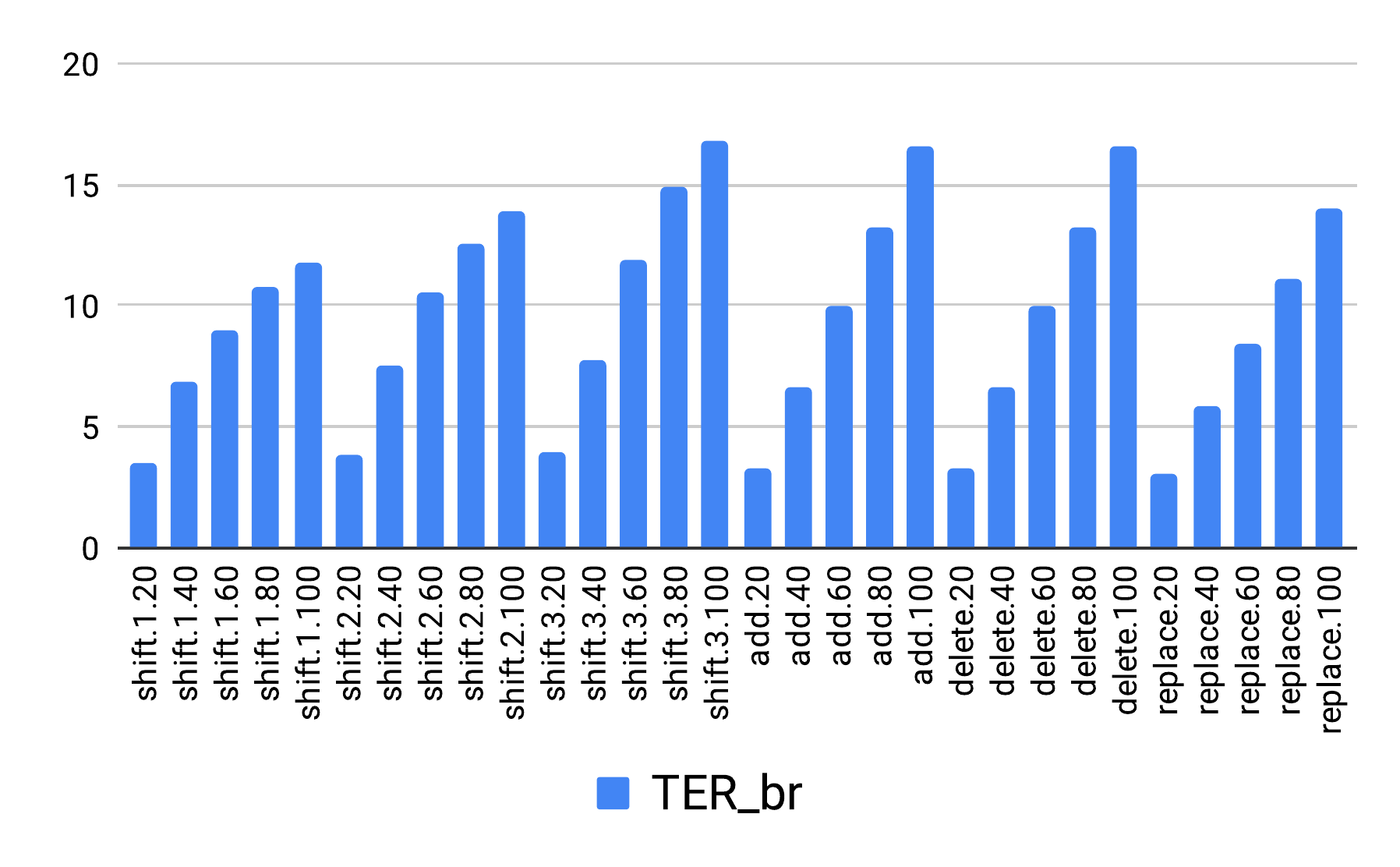}
        %\caption{...} \label{fig:e}
    \end{subfigure}
\caption{Behaviour of segmentation metrics when gradually transforming the reference segmentation with shift.1, shift.2, shift.3, add, delete, and replace operations applied on the boundaries. Metrics that do not distinguish different boundary types were not computed in the ``replace'' scenario.}
\label{fig:metrics_sensitivity}
\end{figure*}

\subsection{Data and implementation}
The subtitling data we use in the experiments come from the MuST-Cinema corpus \cite{karakanta-etal-2020-must}. The test set is compiled from the subtitle files of 9 TED talks, amounting to 545 sentences with subtitle boundaries marked as special symbols. For experiments~\ref{exp1} and \ref{exp2} we use the English side of the English-French pair, while for the boundary projection method we use the French side.

Our code for computing the segmentation metrics is implemented in Python, based on existing libraries. The window-based metrics (\pk{}, WindowDiff), as well as Segment and Boundary Similarity, are computed using the SegEval package\footnote{\url{https://pypi.org/project/segeval/}} \cite{fournier-2013-evaluating}. BLEU and TER are computed with SacreBLEU\footnote{\texttt{BLEU|\#:1|c:mixed|e:no|tok:13a|s:exp|} \texttt{v:2.0.0} \\ \texttt{TER|\#:1|c:lc|t:tercom|nr:no|pn:yes|as:no|} \texttt{v:2.0.0}} \cite{post-2018-call}. 
%%BLEU+\#:1+c:mixed+e:no+tok:13a+s:exp+v:2.0.0TER+\#:1+c:lc+t:tercom+nr:no+pn:yes+as:no+v:2.0.0}
%

\section{Results}
\subsection{How robust/sensitive are metrics to segmentation changes?}

\begin{figure}[!ht]
  \center
  \includegraphics[width=0.9\columnwidth]{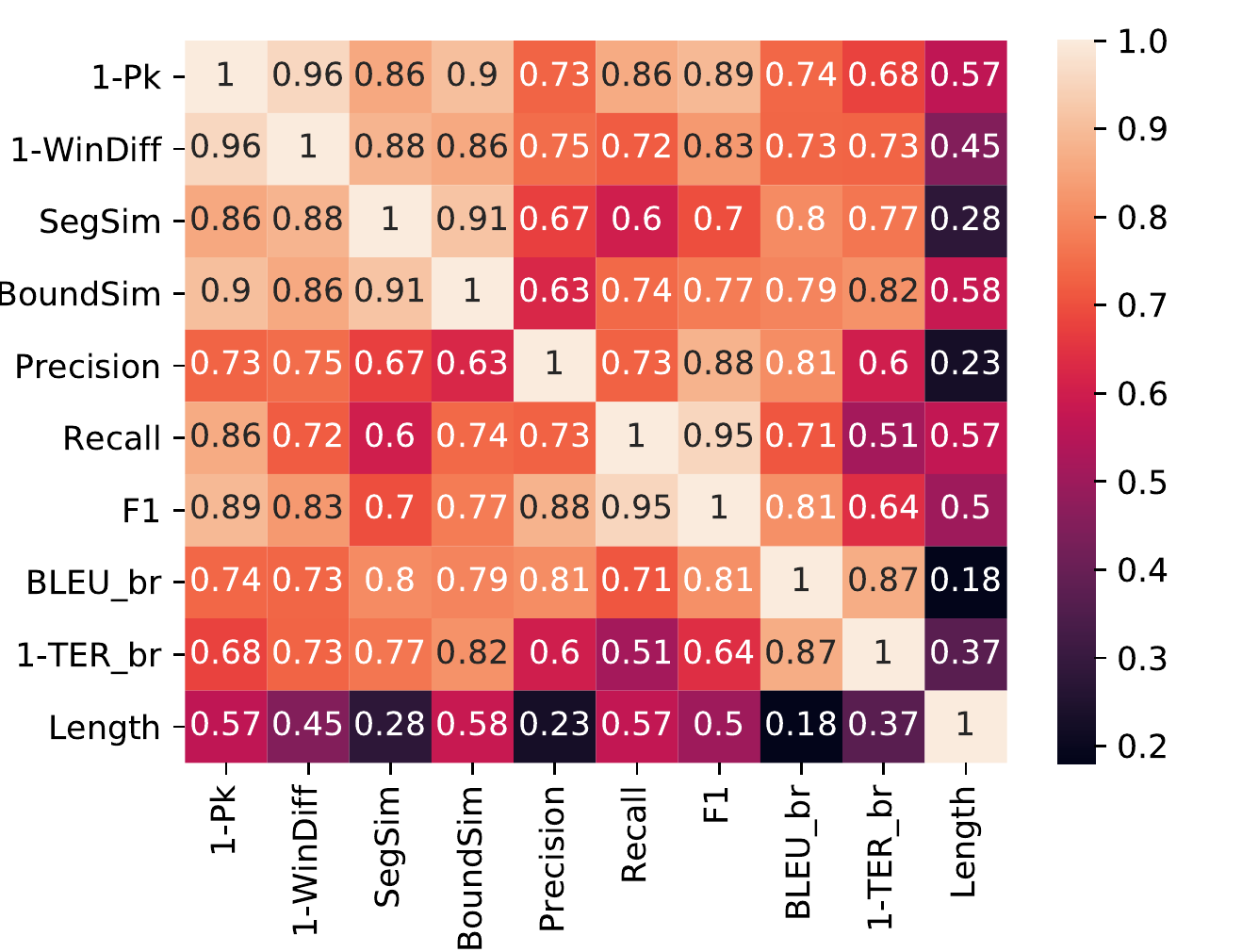}
  \caption{Pearson correlation matrix for the segmentation metrics. Coefficients were computed over the values measured in Experiment 1. ``Length'' is the percentage of lines conforming to the length constraint of max. 42 characters per line. \label{fig:metrics_pcc}}
\end{figure}

We evaluate here the impact of several types and levels of segmentation noise on the segmentation metrics of Section~\ref{exp1}. Apart from the desiderata for a good segmentation metric in Section~\ref{subsec:problem}, the types of noise we apply can have an impact on user experience: Shifts correspond to having the `correct' number of subtitles but segmented in a non-optimal way for comprehension. %We expect that missing the boundary placement by 1 position is more likely to break a syntactic unit than a miss of larger size (the contrary would mean that the majority of words are unitary syntagmas). 
Moreover, near misses are less likely to match reference boundaries, since subtitles rarely contain only 1 token (in our reference, subtitles contain on average 5 tokens\footnote{We acknowledge that this number may vary for languages with different scripts and subtitling conventions.}). Therefore, a shift of 3 positions is more likely to move the boundary in a position where another boundary is placed. For this reason, an optimal metric for subtitling segmentation should not be sensitive to shifting distance by penalising near misses less. As for additions, deletions and replacements, all of them may lead to critical errors; a deletion will lead to overly long subtitles, an addition to shorter subtitles and, as with replacements, to multiple lines in the case of consecutive \eol{}. Since there are no studies clearly showing the effect of each operation on user experience% OR: a reference-based segmentation metric relying on text only cannot be directly linked to user experience?
, we prefer a metric which would equally penalise over- and under-generation of boundaries as well as the generation of the wrong type of boundary. Results for the degradation scenarios are shown in Figure~\ref{fig:metrics_sensitivity}. Correlation between the metrics is analysed through Pearson correlation coefficients (PCC), shown in Figure~\ref{fig:metrics_pcc}.

For precision-recall, shifting the boundaries causes the highest drops, despite the fact that the same number of boundaries is preserved. Interestingly, shifts by 1 position are worse than 2 positions, and in turn worse than 3 positions. The difference is more visible for percentages above 60\%. F1 deteriorates more for deletions than for additions because of a stronger drop in recall. 

By design, the error measured by \pk{} is always lower than that measured by WindowDiff (an error for \pk{} is an error for WindowDiff, but the reverse is not true).
As noted and criticised by \newcite{pevzner-hearst-2002-critique}, \pk{} penalises false negatives (FNs) heavier than false positives (FPs) (in our experiment FNs correspond to deletions, and FPs to additions). 
Thus \pk{} appears to be more recall-oriented than the other metrics, which is confirmed by the higher PCC with the recall metric. However, for \pk{} and WindowDiff penalties increase regularly with shift size and reversely to precision and recall (1$<$2$<$3). Again, these metrics are more sensitive to deletions than to additions. 

SegSim computes cosmetically high values (as mentioned by \newcite{fournier-2013-evaluating}), which can be inconvenient for interpretation since it lacks sufficient resolution.
The new normalisation introduced for BoundSim notably solves this issue. As with the window-based metrics, SegSim and BoundSim are more sensitive to deletions and additions and give less penalty to near misses (here only for shifts of 1). This can be explained by the fact that shifting a boundary by 1 position is accounted as one transposition, while longer shifts cost one addition and one deletion. However, \terbr{} is robust to the type of error, as it shows a balance between deletions and additions, as well as shifts of 3 positions. All edit-based metrics are less sensitive to replacements. 
%For window-based metrics (\pk{}, WindowDiff), penalties increase regularly with shift size (shift.3 being more penalised than corresponding additions or deletions).
%For edit distance-based metrics (SegSim, BoundSim, \terbr{}), shift.1 is sensibly less penalised than shift.2 and shift.3, which in turn are more penalised than additions and deletions.
%although manual coders often agree upon the bulk of where segment lie, they frequently disagree upon the exact position of boundaries \cite{Artstein08}.

\bleubr{} globally remains within the 45--100 range, since it is the only metric considering textual content. It is hardly sensitive to shift size; shifting 1, 2 or 3 positions yields almost the same scores, but shifts are more penalised than the other types of noise. Despite being a precision-based metric, \bleubr{} here is actually robust to error type, since it equally penalises deletions, additions, as well as replacements.
%It is worth considering that one wrongly omitted boundary decreases the number of $n$-grams matches more than one wrongly added boundary. Removing a token from a reference ``breaks'' four 4-grams, three 3-grams, two bigrams, and one unigram. Adding a wrong token breaks one less $n$-gram for each order.
It is worth considering that one wrongly omitted boundary will affect the $n$-gram precision (for $n>1$), although not as much as one wrongly added boundary.
Therefore, this balance between deletion and addition penalisation could be attributed to the effect of brevity penalty, which decreases the score of segments with missing boundaries.  

To conclude, drawing back to our criteria for a good segmentation metric: the ability to account for different types of boundaries is present in SegSim, BoundSim, \bleubr{}, and \terbr{}. \bleubr{} and \terbr{} can take advantage of multiple human references. As for the balance between formal and structural constraints, even though all metrics highly correlate with each other, their correlation with the length conformity (see ``Length'' in Figure~\ref{fig:metrics_pcc}) is low, with only recall, \pk{} and BoundSim correlating above 0.5. For this, precision-recall metrics could give some insights into the type of error (over- or under-segmentation), but should be computed separately for each type of break and combined. %as also suggested by the higher correlation of recall with length conformity. 
This suggests that there is still a long way to go for a segmentation metric that incorporates formal and structural subtitling constraints. 

Apart from the factors mentioned above, sensitivity to near misses is not a desired property for a subtitle segmentation metric. Unlike window- and edit-based metrics, \bleubr{} is in line with our expectations regarding the shifting distance of boundaries. %Given the regular boundary generation by end-to-end systems,  higher penalisation of shifts. 
It also achieves balance between the additions, deletions and replacements. Therefore, \bleubr{} seems to have properties corresponding to our criteria for a good segmentation metric. Still, our expectations on the relationship of metrics with user experience has to be investigated with user studies and the final decision on a good segmentation metric can be taken only after validating its correlation with human judgements.

\subsection{What does \bleubr{} really measure?}

In spite of its flaws, \bleunb{} remains the go-to metric for MT research and studies on automatic subtitling. %\bleunb{} computes a global evaluation of the content match between the reference and the hypothesis text, averaging modified precision scores for $n$-grams of different orders (typically from $n=1$ to $n=4$). 
As for its relation to segmentation quality, in Section~\ref{exp1}, we showed that for perfect output text (\bleunb{}=100) \bleubr{} indeed ranks segmentation from good to bad (fourth graph of Figure~\ref{fig:metrics_sensitivity}), with the added benefit of yielding a global aggregate value taking into account all types of errors (additions, deletions, replacements). The question we now turn to is the usefulness of \bleubr{} \emph{for imperfect texts}. We would like to independently evaluate quality in content prediction (with \bleunb{}) and quality in segmentation prediction. Can \bleubr{}, as used in previous work, or $Sigma$ (equation~\eqref{eq:sigma}), play that role?

%We first acknowledge that for fixed \bleunb{}, \bleubr{} indeed ranks segmentation from good to bad, equally penalising additions, deletions and replacements, and  shifts; shifts  of different amplitude are treated almost equally. Segmentation errors yield a \bleubr{} range that can go from simple to double in absolute terms; even for small amounts of noise, \bleubr{} can drop by approximately 10-20\%. 

\begin{figure}[h]
    \centering
    \includegraphics[width=0.7\columnwidth]{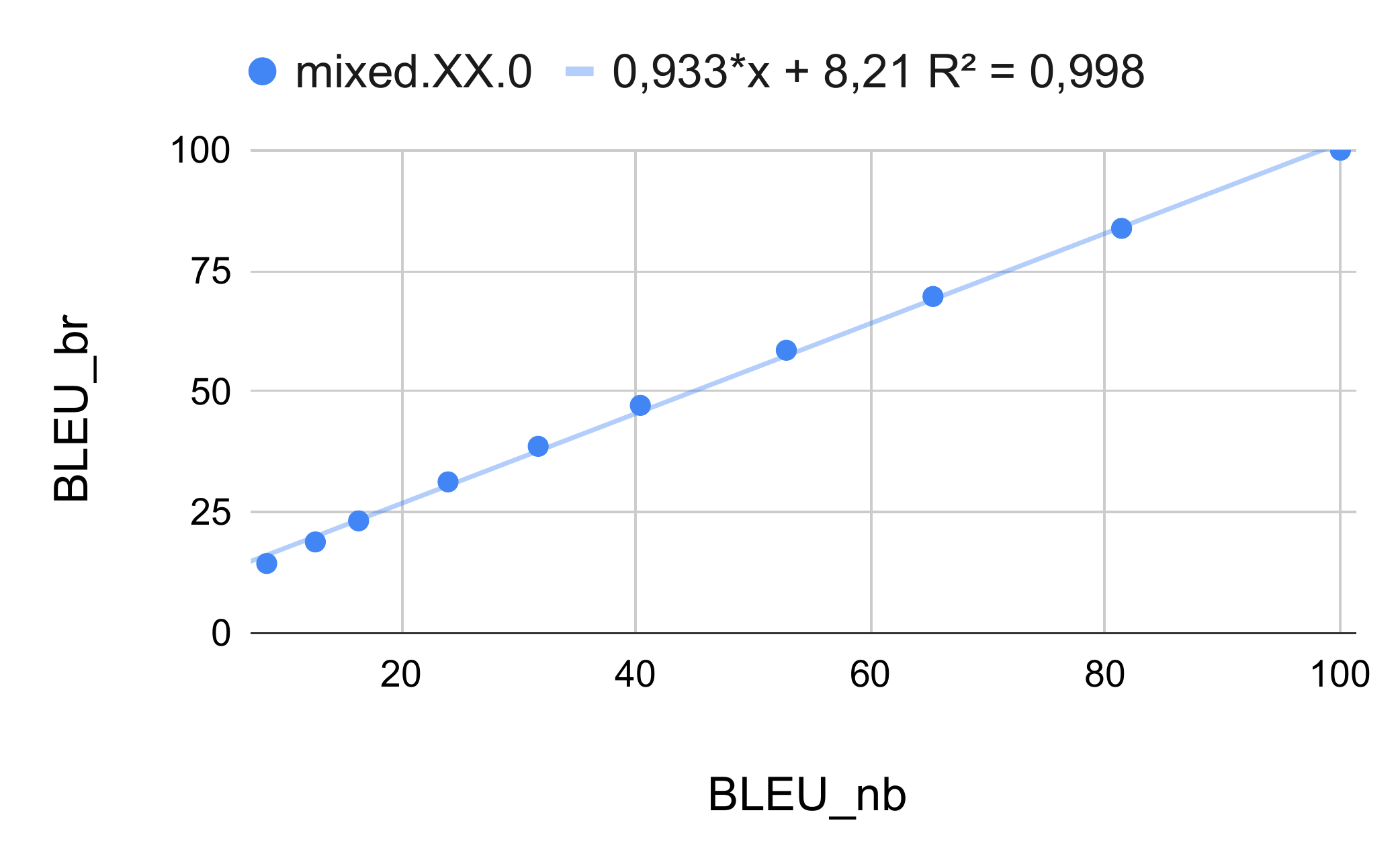}
    \caption{Linearity of \bleubr{} wrt \bleunb{}, for instances where only text was submitted to noise (cf. Section~\ref{exp2}). Linear regression gives a coefficient of determination of 0.998, and a standard error of 1.2.}
    \label{fig:bleu_bleu-br}
\end{figure}

First, for `perfect' segmentation but imperfect text, Figure~\ref{fig:bleu_bleu-br} shows that the relationship between \bleubr{} and \bleunb{} is linear. This indeed confirms our hypothesis that the two metrics are so correlated that their difference cannot be a strong signal of segmentation quality. It also shows that \bleubr{} cannot exceed an upper bound which is strongly related to \bleunb{}, since in this setting the segmentation has not been affected by noise. Given this  \bleubr{}-\bleunb{} dependency, reporting both scores is uninformative and the segmentation signal should be sought in a different relationship.

\begin{figure*}[t]
\centering
	\begin{subfigure}[b]{0.4\textwidth}
        \includegraphics[width=\textwidth]{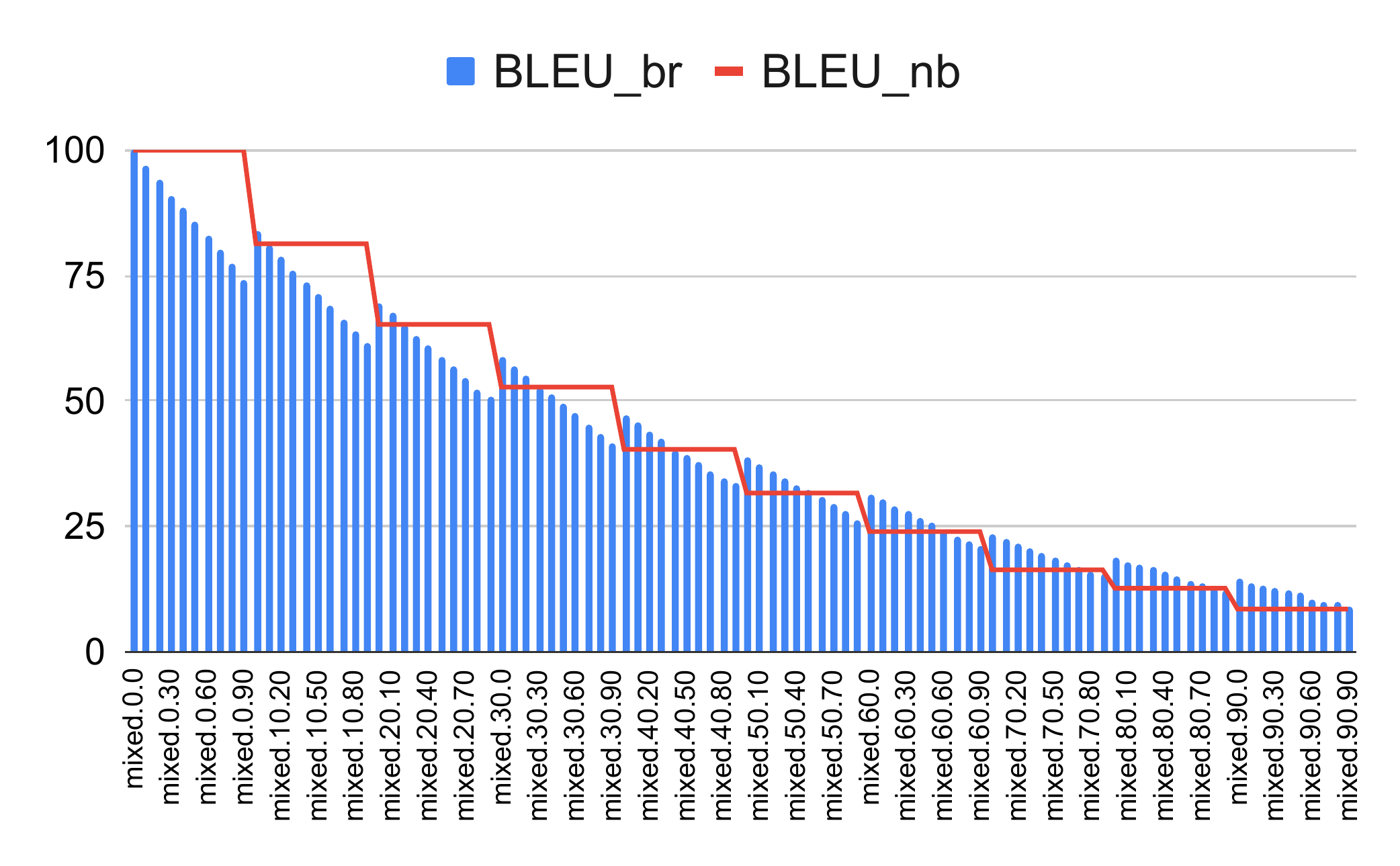}
        \caption{} \label{fig:bleu2}
    \end{subfigure}
    \begin{subfigure}[b]{0.4\textwidth}
        \includegraphics[width=\textwidth]{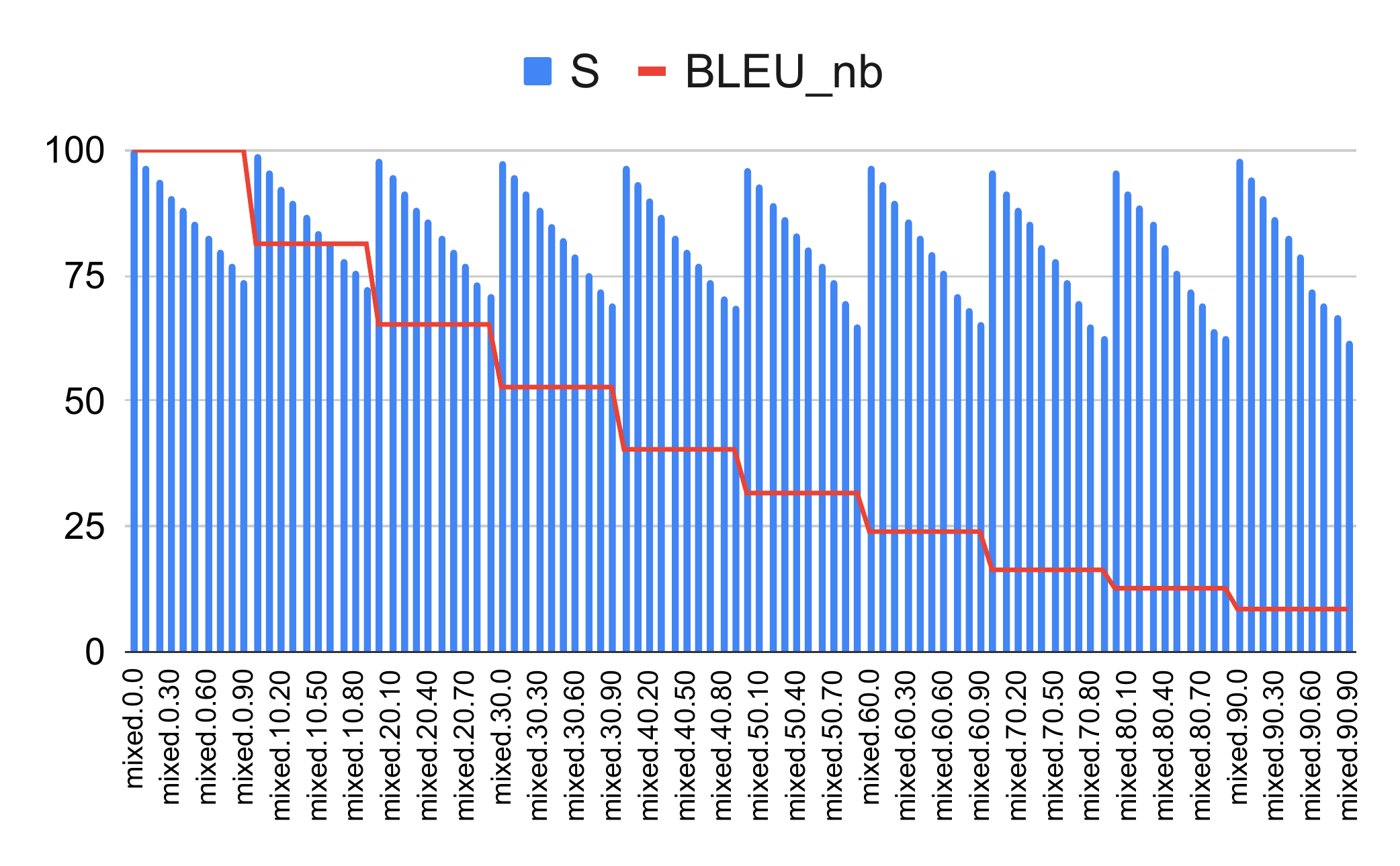}
        \caption{} \label{fig:S}
    \end{subfigure}
    \caption{Values of \bleubr{} (a) and $Sigma$ (b) after applying segmentation noise at different values of \bleunb{} (10 levels of segmentation noise for each \bleunb{} value, 10 \bleunb{} values from 100 down to 8.5). In (a) we see \bleubr{} decreasing with decreasing \bleunb{}, whereas (b) $Sigma$ scores remain stable.
    }
\label{fig:noising}
\end{figure*}

When moving to the scenario of imperfect text and imperfect segmentation, Figure~\ref{fig:bleu2} confirms our hypothesis that the more the quality of \bleunb{} drops, the more often \bleubr{} exceeds \bleunb{}. On the contrary, $Sigma$ remains in a similar range irrespective of the value of \bleunb{}. 
We illustrate this in Figure~\ref{fig:S} where we plot $Sigma$ for various values of \bleunb{}. 
$Sigma$ responds linearly to the amount of segmentation noise, but we observe a minor drift of the range of values when \bleunb{} decreases (from [74.3--100] for mixed.0 to [63.2-95.8] for mixed.90). However, in a realistic scenario, \bleunb{} would typically be constrained to an interval between 25 and 55
% mixed.30 and mixed.60
(corresponding to the 4 central series on Figure~\ref{fig:S}). Moreover, when comparing the segmentation ability of two systems close in \bleunb{} value, the impact of that drift would be all the more limited.
This shows that $Sigma$ can be a good approximate value for capturing segmentation quality, irrespective of the quality of the generated content.

\subsection{Boundary projection} \label{subsec:res3}
We here move to the evaluation of the segmentation of actual outputs from end-to-end systems. Since standard metrics cannot be applied to imperfect texts, we use the boundary projection method as a proxy to compare the system ranking based on all metrics to that of $Sigma$. 
Scores are reported in Table~\ref{tab:MWER-res}. $Sigma$ ranks the output of \texttt{NMT} as best with a score of 89.2, followed by \texttt{Cas} with 83.1, while the two direct systems score very closely with 81.8 for \texttt{e2e$_{base}$} and 81.5 for \texttt{e2e$_{pt}$}. In relation to the standard metrics on the projected reference (columns 2-10) and the metrics considering text quality (\bleubr{} and \terbr{} to the left of $Sigma$), all metrics clearly agree with $Sigma$ on their rating of the output of the \texttt{NMT} system as having the best segmentation among the examined outputs. In TED talks, subtitlers create the target subtitles (our reference) using the source subtitles as template, therefore it is expected that a system receiving the source boundaries as input and able to correctly copy the boundaries will achieve high similarity with the reference. However, when comparing the scores for the three Speech Translation systems (\texttt{Cas}, \texttt{e2e$_{base}$} and \texttt{e2e$_{pt}$}), the agreement among metrics is lost. The cascade output seems to have better segmentation with 6 wins (WindowDiff, SegSim, precision, \bleubr{} and the two versions of \terbr{}) and 3 ties (\pk{}, F1 and \bleubr{} on system output) over the \texttt{e2e$_{pt}$}, which is ranked best according to BoundSim and recall. The close scores in many metrics, including $Sigma$, show no clear winner between the two direct systems. This is expected since \texttt{e2e$_{pt}$} was pre-trained on non-segmented text, which improves translation quality, but did not receive any additional segmentation data compared to \texttt{e2e$_{base}$}. The length of the predicted output seems to have an effect, since metrics with a high correlation with length conformity rank \texttt{e2e$_{pt}$} higher than \texttt{Cas} (length conformity is 95\% for \texttt{e2e$_{pt}$} and 91\% for \texttt{Cas}). All in all, the two ST systems seem to be making different types of errors (\texttt{Cas} has higher precision, \texttt{e2e$_{pt}$} higher recall), but for most metrics the scores are so close that it may be hard to tell which output is the best. %Between the two e2e outputs, \texttt{e2e$_{pt}$} scores as better or at least equal with the base model, except for \bleubr{} where the difference is not significant ($p=.305$ as computed using the paired bootstrap resampling).

%When comparing \bleubr{} and \terbr{} computed between the projected reference and the true reference with the \bleubr{} and \terbr{} between the system output and the true reference  in Table~\ref{tab:MWER-res}), we observe that the ranking of the outputs differs too. The \bleubr{} on the projected reference ranks \texttt{Cas} as best, while \bleubr{} computed on the system output does not distinguish between \texttt{Cas} and \texttt{e2e$_{pt}$}. This is another indication that \bleubr{} is influenced by text quality. The boundary projection method, even though possibly containing alignment errors, allows us to disentangle the effect of text quality from segmentation in the case of \bleubr{}. %The same effect is observed for \terbr{}. Even though \terbr{} is computed by masking all words except boundaries, this masking is not enough to totally eliminate the effect of text quality on segmentation. However, because of its biases due to shift size as shown in Experiment~\ref{exp1}, \terbr{} may not be an optimal metric for evaluating segmentation for perfect texts.
%Therefore, in the absence of a metric able to distinguish the effect of good segmentation between outputs of varying content quality, our boundary projection method is one possible solution for evaluating segmentation with standard metrics.

The results show that, even though the metrics are capable of properly rewarding high quality output, distinguishing between outputs of similar segmentation quality under real evaluation settings is a difficult task and requires metrics with sufficient resolution. Despite this, $Sigma$ shows a relatively high agreement with the majority of metrics in ranking the cascade output as best among the ST outputs. The boundary projection method is used here as a proxy for disentangling the effect of text quality from segmentation, but the scores computed through this method are impacted by the performance of the alignment algorithm, especially for low quality outputs. On the other hand, metrics computed directly on imperfect texts (\bleubr{} and \terbr{}) are strongly influenced by translation quality, as shown by the different ranking of \bleubr{} computed on projected reference (column 9) vs. system output (column 11). $Sigma$ is not constrained by either of these limitations and provides a clear, interpretable and easy-to-compute solution for evaluating the segmentation even between imperfect texts of similar quality.

\begin{table*}[t]
\small
    \centering
    \begin{tabular}{l|ccccccccc|cc|c} \toprule
       System & \pk{} & Windiff & SegSim & BndSim & Prec & Rec & F1 & \bleubr{} & \terbr{} &  \bleubr{} & \terbr{} & \textbf{$S$} \\ \midrule 
    NMT & .192 & .208 & .979 & .637 & .711 & .735 & .723 & 83.18 & 6.87 & 32.16 & 19.38 & 89.2 \\%89 \\
    Cas  & \underline{.252}	& \underline{.270}	& 	\underline{.970} &	.519 &	\underline{.639}	& .667	& \underline{.653}	& \underline{76.14}	& \underline{8.91}	& 26.34 & \underline{23.23} & \underline{83.1}\\%83 \\
    e2e$_{base}$ & .257 &	.277 &	.969 &	.515 &	.601 & .667 &	.632 & 75.00 & 9.29 & 22.53 & 24.48 & 81.8\\%81 \\
    e2e$_{pt}$ & \underline{.252} & .276 & .969 & \underline{.525} & .610 & \underline{.702} & \underline{.653} & 74.89 & 9.24 & \underline{26.36} & 23.52 & 81.5 \\ \bottomrule %84
    \end{tabular}
    \caption{Segmentation scores of $Sigma$ and the examined metrics for the output of four systems after projecting their boundaries to the reference (Section~\ref{subsec:res3}). The two columns to the left of $Sigma$ are \bleubr{} and \terbr{} scores between the output and the reference without projection% reported in Karakanta et al. (2020b)
    . Best score among the ST outputs is underlined.
}
    \label{tab:MWER-res}
\end{table*}

\section{Related work\label{sec:related}}
Automatic subtitle segmentation has been previously evaluated in the case of interlingual and intralingual subtitles, by comparing the automatically generated output against a reference \cite{alvarez-2014-customised}. For interlingual subtitles, \newcite{alvarez-et-al-2016} proposed a segmentation algorithm based on Logistic regression and Support Vector Machine classifiers. The evaluation was performed with precision-recall-F1 measures, based on the ability of the algorithm to insert a segmentation boundary, without however distinguishing between line and subtitle breaks. Later, \newcite{Alvarez17improving}
compared rule-based and machine-learning segmentation methods with metrics which considered either subtitle breaks only or both line and subtitle breaks. These metrics count the segmentation errors (F1, NIST), the number of incorrectly segmented portions (DSER) or the edit distance between sequences of reference positions and hypothesis positions (SegER). \newcite{karakanta-etal-2020-point} trained a sequence-to-sequence model on different combinations of real and synthetically segmented data, which transforms an unsegmented sentence into a sentence with line and subtitle breaks. Except for F1, performance is evaluated with BLEU as a similarity measure, and characters per line (CPL), as the percentage of segmented subtitles conforming to the length constraint.

Segmentation has also been evaluated in the context of Machine Translation for subtitling.  \newcite{Karakanta20is42} report BLEU$_{br}$, where the BLEU metric of \newcite{Papineni02bleu} is computed on text containing subtitle boundaries annotated as special symbols. Each boundary symbol counts as one token in the BLEU computation. Similarly, \newcite{Matusov19customizing} report scores for their MT metrics in the \textit{S-mode}, where line breaks in a subtitle are marked with a separator symbol. However, in their setting there is a one-to-one correspondence between source-target subtitles. \newcite{Karakanta20is42} also introduce TER$_{br}$, a variant of TER \cite{Snover06study} which is computed on text where all words except for the subtitle boundaries are masked. The authors claim that this metric determines the effort required by a human subtitler to manually correct the segmentation, ignoring word errors. Last, \newcite{cherry-21} propose two metrics again related to BLEU: 1) \emph{Timed-BLEU}, where target to reference alignments necessary for the evaluation are created by linear temporal alignment, over which BLEU is calculated as usual, and 2) \emph{T-BLEU Headroom}, calculated as the difference between an upper bound of T-BLEU and the actual T-BLEU. % lower-is-better boundary error rate, interpretable as the amount of T-BLEU that could be recovered by improving only the position of the boundaries. 
Both metrics only apply when the output contains timestamps, which are not always available in subtitle generation with end-to-end systems.

\section{Conclusion}\label{sec:conclusion}
We have analysed metrics and methods to evaluate the segmentation of text into subtitles, given a human reference. Our analysis using artificial noise in segmentation has shown that for perfect texts, \bleubr{} satisfies our criteria for a good subtitle segmentation metric. However, when moving to imperfect texts, \bleubr{} correlates highly with regular BLEU, therefore the segmentation signal cannot be extracted by a simple difference between \bleubr{} and \bleunb{}. We thus introduce a new subtitle segmentation score $Sigma$, as the ratio of \bleubr{} to its approximated upper bound. %This allows to distinguish the signal of good segmentation irrespective of text quality. 
In order to compare $Sigma$ with standard segmentation metrics for evaluating real system outputs, we further proposed a boundary projection method which projects the subtitle boundaries from the output to the reference. We noted that in real evaluation settings existing metrics do not always agree on their ranking of the outputs, especially for outputs of similar quality. 
We believe that the final response on the most accurate method to evaluate subtitle segmentation can be given only after obtaining correlations of the metrics and methods proposed in this paper with human judgements. However, the analysis presented in this work has shed light into the critical aspects of evaluating subtitle segmentation, in order to better design user studies to collect these human judgements and to refine the approximation of upper bound \bleubr{} for computing $Sigma$ in subtitling segmentation tasks.

\section{Acknowledgements}
This work was partially supported by the European Commission funded project ``Humane AI: Toward AI Systems That Augment and Empower Humans by Understanding Us, our Society and the World Around Us'' (grant \#820437) and by the BPI-France investment programme "Grands défis du numérique", as part of the ROSETTA-2 project (Subtitling RObot and Adapted Translation). The support is gratefully acknowledged.

%\section*{Appendix: How to Produce the \texttt{.pdf}}

% \nocite{*}
\section{Bibliographical References}\label{reference}
%\label{main:ref}

\bibliographystyle{lrec2022-bib}
\bibliography{lrec2022-example}

%\section{Language Resource References}
%\label{lr:ref}
%\bibliographystylelanguageresource{lrec2022-bib}
%\bibliographylanguageresource{languageresource}

\end{document}